\documentclass{article}

\usepackage{PRIMEarxiv}

\usepackage[utf8]{inputenc} 
\usepackage[T1]{fontenc}    
\usepackage{hyperref}       
\usepackage{url}            
\usepackage{booktabs}       
\usepackage{amsfonts}       
\usepackage{nicefrac}       
\usepackage{microtype}      
\usepackage{lipsum}
\usepackage{fancyhdr}       
\usepackage{graphicx}       
\usepackage{mathrsfs}
\usepackage{dutchcal}
\usepackage{amsmath,amssymb,amsfonts}
\graphicspath{{media/}}     

\title{Efficient Self-Supervised Grading of Prostate Cancer Pathology
}

\author{
  Riddhasree Bhattacharyya \\
  Machine Intelligence Unit,\\ 
  Indian Statistical Institute,\\ 
  Kolkata 700108, India \\
  \texttt{riddhasreeb@gmail.com} \\
   \And
  Surochita Pal Das \\
  Machine Intelligence Unit,\\ 
  Indian Statistical Institute,\\ 
  Kolkata 700108, India \\
  \texttt{pal.surochita@gmail.com} \\
  \And
  Sushmita Mitra \\
  Machine Intelligence Unit,\\ 
  Indian Statistical Institute,\\ 
  Kolkata 700108, India \\
  \texttt{sushmita@isical.ac.in} \\
}

\begin{document}
\maketitle

\begin{abstract}
Prostate cancer grading using the ISUP system (International Society of Urological Pathology) for treatment decisions is highly subjective and requires considerable expertise. Despite advances in computer-aided diagnosis systems, few have handled efficient ISUP grading on Whole Slide Images (WSIs) of prostate biopsies based only on slide-level labels. Some of the general challenges include managing gigapixel WSIs, obtaining patch-level annotations, and dealing with stain variability across centers. One of the main task-specific challenges faced by deep learning in ISUP grading, is the learning of patch-level features of Gleason patterns (GPs) based only on their slide labels. In this scenario, an efficient framework for ISUP grading is developed.  

The proposed TSOR is based on a novel {\bf T}ask-specific {\bf S}elf-supervised learning (SSL) model, which is fine-tuned using
{\bf O}rdinal {\bf R}egression. Since the diversity of training samples plays a crucial role in SSL, a patch-level dataset is created to be relatively balanced w.r.t. the Gleason grades (GGs). This balanced dataset is used for pre-training, so that the model can effectively learn stain-agnostic features of the GP for better generalization. In medical image grading, it is desirable that misclassifications be as close as possible to the actual grade. From this perspective, the model is then fine-tuned for the task of ISUP grading using an ordinal regression-based approach. Experimental results on the most extensive multicenter prostate biopsies dataset (PANDA challenge), as well as the SICAP dataset, demonstrate the effectiveness of this novel framework compared to state-of-the-art methods.
\end{abstract}


\section{Introduction}
\begin{figure}[!ht]
    \centering
    \includegraphics[width=0.7\textwidth]{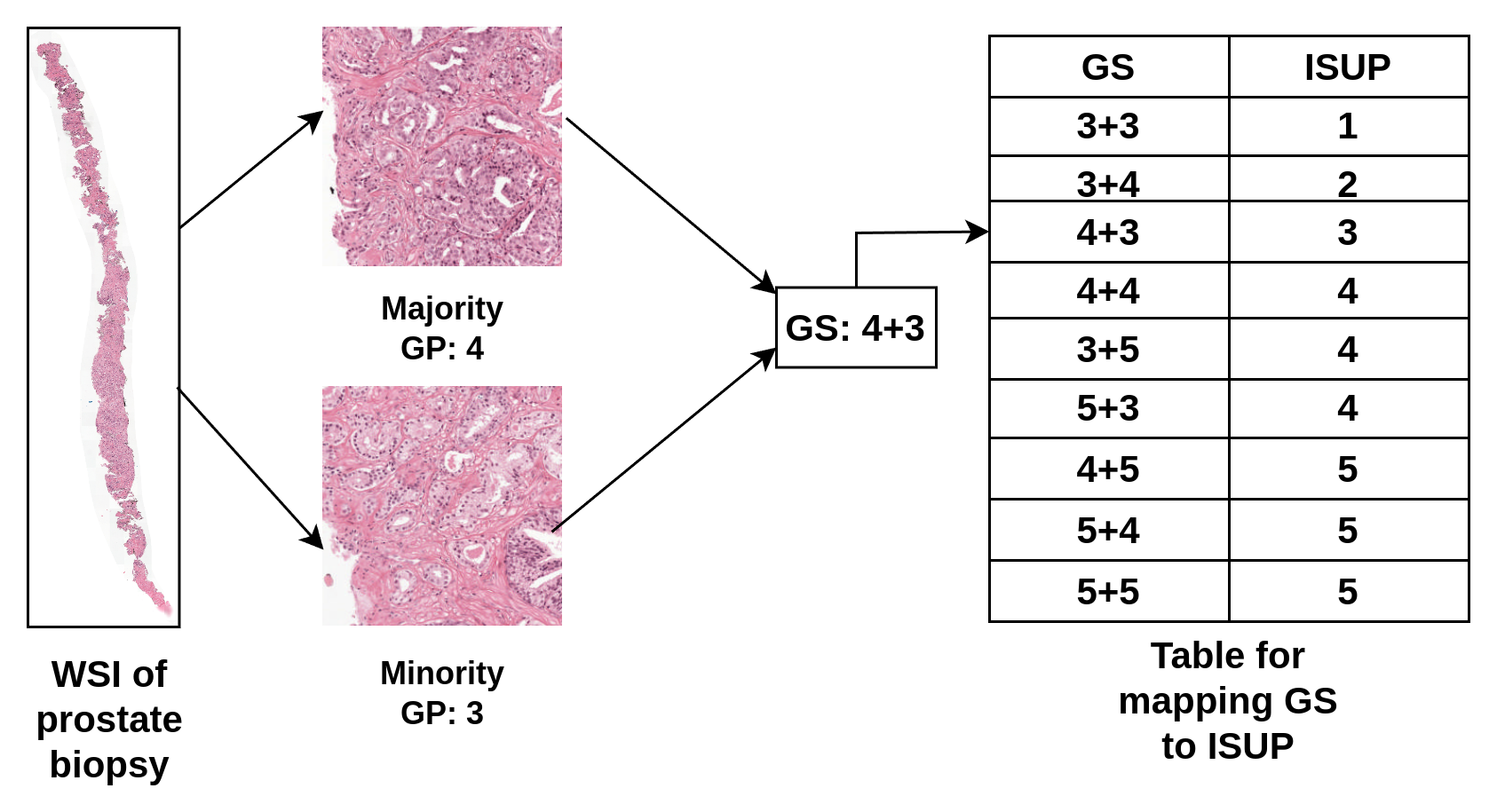}
    \caption{The ISUP grading system, based on Gleason Pattern (GP) and corresponding Gleason Score (GS)}
    \label{isup}
\end{figure}

{P}{rostate} cancer is the second most common cancer that affects men worldwide \cite{yang2021multi}. The gold standard for diagnosing and grading it, for determining the prognosis, typically involves Gleason grading of biopsies. A general workflow of pathologists for grading prostate biopsies includes the collection of a small amount of prostate tissue, laminating, staining with Hematoxylin and Eosin (H\&E), followed by its staging. The Gleason grading system, with an inverse correlation between the degree of tissue gland differentiation and the GG, contains three values ranging from 3 to 5 \cite{silva2020going}. The new ISUP grading system \cite{epstein2015new}, as shown in Fig. \ref{isup}, considers the first and second most frequent GP, in terms of proportion and severity within a prostate biopsy, followed by mapping it to an ISUP grade (1-5). Benign biopsies are assigned an ISUP grade of 0. 

The entire procedure for grading prostate biopsies by pathologists is cumbersome, requires considerable expertise, and is highly prone to intraobserver and interobserver variability. This highlights the need for computer-assisted prostate cancer diagnosis/ progression systems \cite{bulten2020automated}. With the advent of digitized biopsies, the increased availability of Graphic Processing Units (GPUs) at reasonable cost, and large publicly available datasets, the development of deep learning automated systems in histopathology has increased manifold \cite{liu2022deep,li2020deep,qu2022towards}. 
Deep learning \cite{lecun15}, particularly Convolutional Neural networks (CNN), work directly on the pixel values of an input image. Hence, it overcomes manual errors caused by inaccurate segmentation and/or hand-crafted feature extraction of shallow learning.  

As the WSIs contain billions of pixels, it becomes computationally infeasible for CNNs to handle. This requires division of each image into patches. Given the large size of digitized biopsies, it is often impossible to obtain patch-level annotations; thereby, making it difficult to train fully supervised deep learning models at patch level. Thus, the huge and diverse data of WSIs required by supervised algorithms, combined with the limited availability of expert pathologists, resulted in a paradigm shift to weakly supervised and self-supervised learning.

 Weakly supervised learning, including Multiple Instance Learning (MIL) \cite{gadermayr2024multiple}, predicts the grade of a WSI based on slide-level labels which are readily available from clinical records. The use of MIL in histopathology includes instance- and bag-based approaches, the former being more popular due to its robustness \cite{qu2022towards}. Instance-based MIL classifiers are trained on patch labels inferred from WSI labels, with MIL pooling for WSI-level prediction. Bag-based MIL classifiers combine patch embeddings to obtain a WSI-level representation for training with WSI labels. 
 Multiple scale patch embeddings were aggregated, using Dual Stream MIL, for WSI classification in DSMIL \cite{li2021dual}.
Instance-level clustering of informative patches enabled CLAM \cite{lu2021data} to refine the feature space while predicting WSI labels. Contrastive learning over WSI helped improve feature aggregation in SCL-WC \cite{wang2022scl}, by enhancing interclass separability and intraclass compactness.  TransMIL \cite{shao2021transmil} used a transformer to capture morphological and spatial information from patches, for weakly supervised WSI classification.

Self-Supervised Learning (SSL) aims to reduce the manual annotation burden by learning effective feature representations, from a set of unlabeled data. This process of feature learning, through some auxiliary tasks, is termed pre-training. It can be task-specific or task-agnostic, depending on whether or not the training samples and the auxiliary task are considered to be solely based on the target task. The pre-trained model is then fine-tuned for the target task based on a limited amount of labeled data. Recent research demonstrates the effectiveness of the SSL paradigm in histopathology \cite{chen2020simple, qu2022towards, haghighi2024self}.

Patch-level Gleason grading of WSIs was explored in Ref. \cite{silva2020going, silva2021self}. However, patch-level annotations are difficult to obtain. Hence, research now focuses on WSI-level Gleason grading or ISUP grading, without involving patch-level annotations. A supervised patch-level classifier was initially trained to predict GGs with patch-level annotations \cite{nagpal2019development,silva2020going}, and utilizing patch labels inferred from WSI labels \cite{silva2021self}. The predicted patch-level grades were converted into features (indicative of the amount of GP in a WSI) to determine the corresponding WSI grade by the $k$-nearest neighbors ($k$NN) classifier \cite{nagpal2019development}. The average patch-level features, extracted from the trained classifier, were used for the WSI-level grading by a $k$NN and Multilayer Perceptron (MLP) in Ref. \cite{silva2021self}. The GG percentages in each WSI, determined from the biopsy-level prediction map (constructed from patch-level predictions), were used to calculate WSI-level grades using MLP \cite{silva2020going}.  

The bag-based MIL approach, with different variations, was explored for ISUP grading in Ref. \cite{nirthika2020loss,yang2021multi,yang2023devil}. The Mean Squared Error loss  was used in conjunction with the bag-based MIL approach in \cite{nirthika2020loss}. The extracted features for ISUP grading were refined by incorporating Multichannel and Multispatial Attention in a CNN \cite{yang2021multi}. The ISMIL framework \cite{yang2023devil}, on the other hand, improved the recognition of small tumor regions by intensively sampling crucial regions.

\subsection{Motivation}
\label{intr:motivation}

An ISUP grade prediction model must learn to effectively discriminate between the features of different GPs. This is best possible with training at the patch level \cite{epstein2015new,silva2020going}. However, when only 
slide-level labels (with regard to primary and secondary GPs of WSIs) are available, it becomes difficult to infer the labels of individual patches \cite{qu2022towards}. Hence, the SSL paradigm seems to be the preferred choice for learning features of GPs from unlabeled patches.

The diversity of training samples plays a crucial role in SSL. It becomes computationally expensive to train a task-agnostic SSL framework using a large database of histopathological images of different types of cancer. Here, learning can be effective only for some general characteristics of the images. Cancer patterns are typically heterogeneous and unique in prostate cancer images. Here, it becomes desirable to use a task-specific SSL framework with prostate cancer-specific data. Again, randomly sampling patches from the prostate WSIs results in large datasets having high class imbalance. Precisely, the number of benign patches becomes maximum, with Gleason grade 5 patches being minimum \cite{silva2020going}. In this scenario, SSL-based models face difficulty in learning discriminating features from the corresponding GP \cite{chen2020simple, qu2022towards}. In order to circumvent this problem, we create a relatively balanced patch-level dataset w.r.t. GG by 
pseudo-labeling of patches with MIL.

The workflow of instance-based MIL involves supervised training of a patch-level classifier, with aggregation of patch-level outputs to generate a WSI-level prediction. As patch-level annotations are unknown, these approaches typically label all patches from a cancerous WSI as cancerous. As it results in noisy training, the patch labels cannot be predicted with high confidence using such frameworks. It is also not suitable for prostate WSIs, where each cancerous WSI can have multiple GPs. Bag-based MIL, on the other hand, entails extracting and combining the features of all patches to generate the WSI-level features. These are used in the supervised training of the bag-level classifier with the WSI labels. Although there is no noise involved during this training, the models are not suitable for the prediction of patch labels \cite{qu2022towards}. Therefore, we adopt a hybrid MIL approach for training our model for pseudo-labeling of patches. Specifically, it involves simultaneous training of instance-based and bag-based classifiers, to balance the trade-off between the two MIL approaches. 

Another practical problem in analyzing histopathological images, collected from different centers, is the inherent variation in staining. As color is a crucial feature for deep neural networks, this often leads to under-performance. Two techniques commonly used to address this issue are stain normalization and stain augmentation. The success of stain normalization is based on the selection of a template image with which the stain-color distribution of the source images can be aligned. However, manual selection of such a template is often difficult \cite{tellez2019quantifying}. Stain augmentation, on the other hand, attempts to simulate different stain variations while preserving the morphological characteristics. This allows the model to learn stain-agnostic characteristics \cite{tellez2018h}. We introduce an additional loss term, based on two different stain-augmented views of each patch, during pre-training of the Task-Specific SSL model. This helps in learning the stain-agnostic morphological characteristics of the GP.

Errors in medical image grading should be as close as possible to the actual grade for a model to be called efficient. Therefore, an ordinal regression-based approach is used to fine-tune the pre-trained network. Thus, the model tries to minimize the distance between the predicted and actual labels; instead of penalizing each misclassification equally.

\subsection{Contribution}

\begin{figure*}[!ht]
    \centering
    \includegraphics[width=0.9\textwidth]{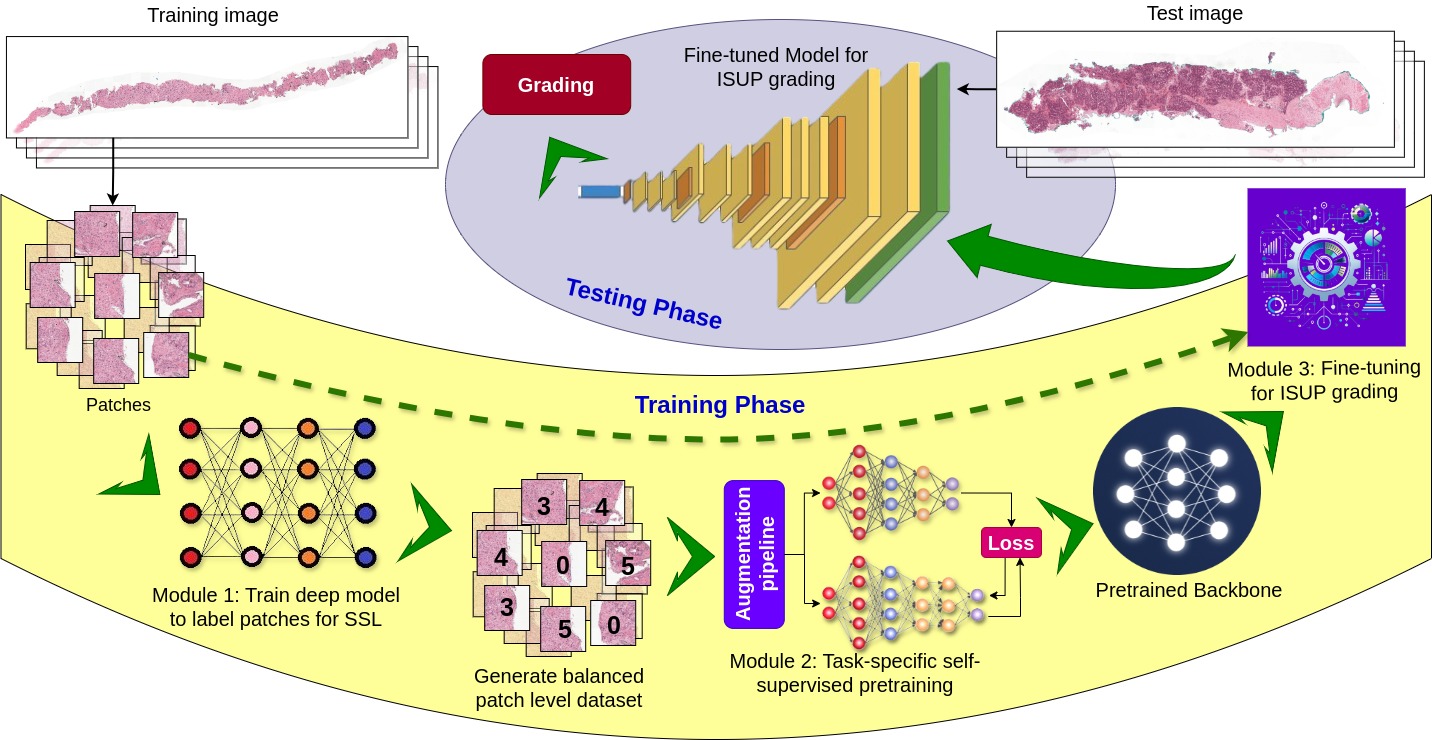}
    \caption{Schematic representation of the proposed framework $TSOR$}
    \label{fig:workflow}
\end{figure*}

A new and efficient SSL-based ISUP grading system is developed for prostate WSI using only
slide-level labels. A patch-level pre-training framework, using a novel task-specific and stain-agnostic SSL, helps to effectively learn the patch-level features of the GPs. The model is then fine-tuned for the task of ISUP grading, using an ordinal regression-based approach. This is unlike the previous approaches where patch-level annotations were needed to learn the patch-level features. Neither is it as computationally expensive as the 
task-agnostic SSL frameworks, typically employed to learn general patch-level features. The framework also tries to address the problem of stain variation while additionally considering the ordinal nature of the classes during fine-tuning, for an effective ISUP grading across different datasets.
The block diagram in Fig. \ref{fig:workflow} outlines the interconnections between the individual modules involved,  in our $TSOR$ framework, with the main contributions summarized below. 
\begin{enumerate} 
\item 
Preparation of patch-level dataset, relatively balanced w.r.t. GGs, for effective self-supervised pre-training. This is achieved through the labeling of the patches in an efficient hybrid MIL framework.

\item
Design of a novel task-specific SSL framework with a new loss term based on stain augmentation. This allows the model to learn the stain-agnostic morphological features of the GPs.

\item Fine-tuning of the pre-trained network, in terms of ordinal regression, for ISUP grading. 

\end{enumerate}

The effectiveness of the model, as compared to related state-of-the-art methods, is established through validation on benchmark PANDA Challenge and SICAP datasets.

The rest of the article is organized as follows. Section \ref{sec:method} introduces the $TSOR$, a new Task-specific SSL framework fine-tuned using Ordinal Regression, for ISUP grading. Section \ref{sec:exp_setup} describes the experiments carried out to evaluate the efficacy of $TSOR$ w.r.t. related state-of-the-art (SOTA) algorithms, along with suitable quantitative and qualitative results. Finally, Section \ref{sec:conclusion} concludes the article. 

\section{Methodology}
\label{sec:method}

Let $\{s_n\}_{n = 1}^N$ be a set of $N$ WSI of prostate biopsies. Each $s_n =\{x_{n,i}\}_{i=1}^l$ is represented as a bag of $l$ patches, where $x_{n,i}$ denotes the $i$th  patch cropped from the $n$th slide. The aim is to develop a model to predict the ISUP grade of a WSI using only slide-level information, as available from clinical records in terms of Gleason grade (GG). This corresponds to the primary GG $z_{n,pg}$, secondary GG $z_{n,sg}$, and ISUP grade $z_{n,isup}$ of each $s_n$. Our $TSOR$ framework achieves this goal by using three distinct modules, as described below.

\subsection*{Module 1: Preparation of training data for SSL}

\begin{figure*}[!ht]
    \centering
    \includegraphics[width=0.9\textwidth]{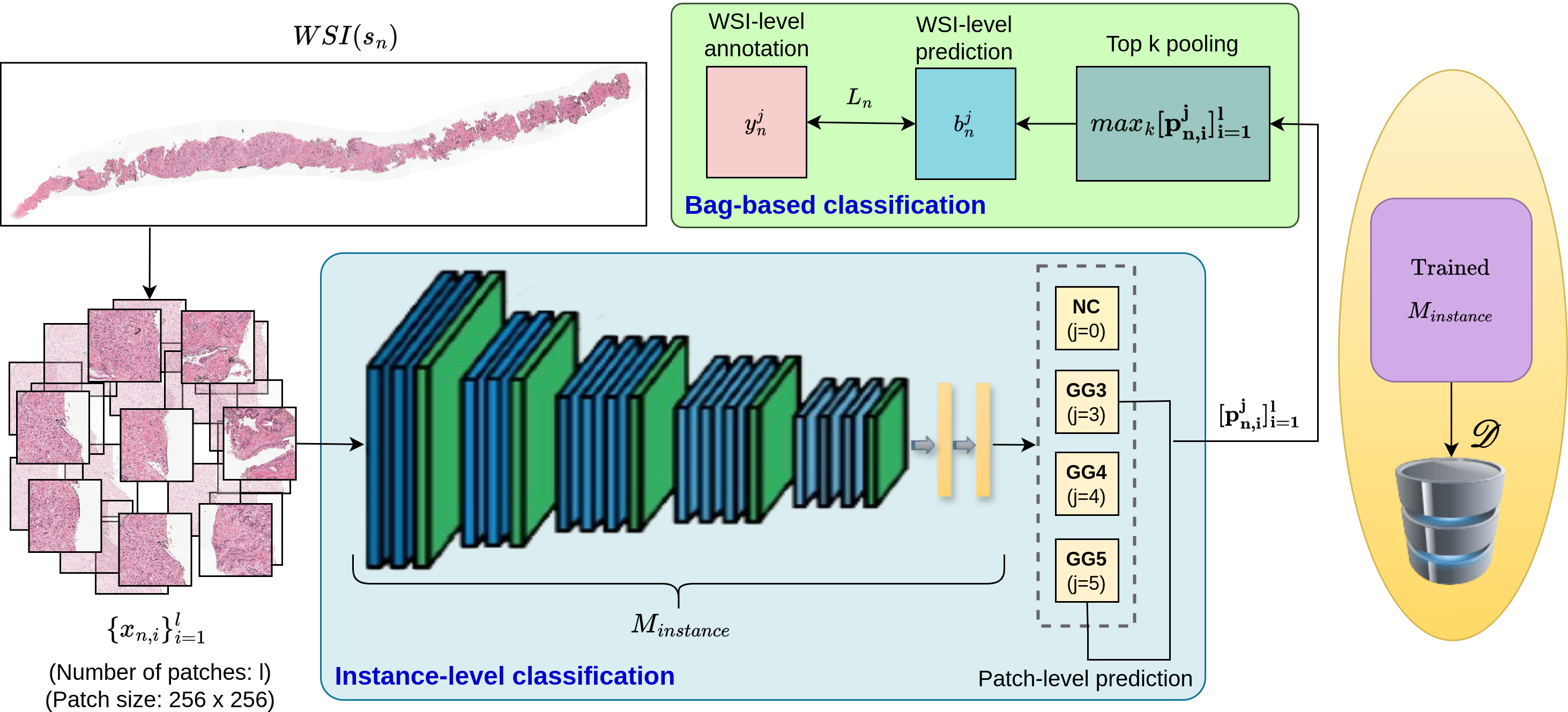}
    \caption{\textbf{Module 1:} Training framework, to create patch level dataset, for SSL}
    \label{fig:step1c}
\end{figure*}

A patch-level data set is created for an effective SSL, in an innovative way such that the dataset is relatively balanced with respect to GGs. A deep learning model, based on the hybrid MIL framework, is trained to label patches using slide-level GGs in an end-to-end fashion. The model training procedure is described in Fig. \ref{fig:step1c}. The first part consists of an instance-level classifier $M_{instance}$ that predicts the probability $p_{n,i}^j$ of a patch $x_{n,i}$ to belong to one of the four classes $j$. Here $j$ = 0 indicates benign, with $j$ ranging from 3 to 5 for $GG3, GG4, GG5$. 

The second part involves bag-based classification, where all patch-level probabilities $p_{n,i}^j$ in each WSI $s_n$ are converted to the corresponding bag-level probabilities $\mathbf{[b_{n}^j]}$ for GG $j=3,..,5$. We have
\begin{equation}
    \mathbf{[b_{n}^j]} = \left[\frac{\sum max_k\mathbf{[p_{n,i}^j]_{i=1}^l}}{k}\right],
\label{avgtopk}
\end{equation}
where each component $b_{n}^j$ represents the average of the corresponding top $k$ (selected by $max_k$) $p_{n,i}^j$ of the WSI. If a WSI has a small amount of GP, then averaging all $p_{n,i}^j$s can reduce the 
WSI-level probability $b_{n}^j$. This might lead to inaccurate predictions. Using a maximum of $p_{n,i}^j$s ensures a correct classification of patches possessing a large amount of GP. In this context, averaging the top $k$ $p_{n,i}^j$ increases the robustness of the model; to also classify the patches containing moderate amounts of GP.

Given that there exist two GGs corresponding to each WSI, we pose this as a multilabel binary classification problem with the loss defined as 
\begin{equation}
    L_{n} = BCE(\mathbf{[b_n^j]}, \mathbf{[y_n^j]}).
\label{ml_bce}
\end{equation}
Here $BCE$ is the Binary Cross Entropy between the predicted bag-based probability at WSI-level of eqn. (\ref{avgtopk}) and the one-hot encoded target output $\mathbf{[y_n^j]}$. The value of each $y_n^j$ is 0 if the WSI does not exhibit GG $j$ (in $z_{n,pg}$, $z_{n,sg}$), and 1 otherwise. 

The trained deep model $M_{instance}$ is then used to label the patches. An approximately equal number of patches are selected from each class to create the patch-level dataset $\mathscr{D}$ for self-supervised pre-training. Patches belonging to each class are initially sorted, based on tissue percentage, followed by the selection of the top $k$ patches. 

\subsection*{Module 2: Task-specific \& stain-agnostic SSL pre-training}
\label{step2}

\begin{figure*}[!ht]
    \centering
    \includegraphics[width=0.9\textwidth]{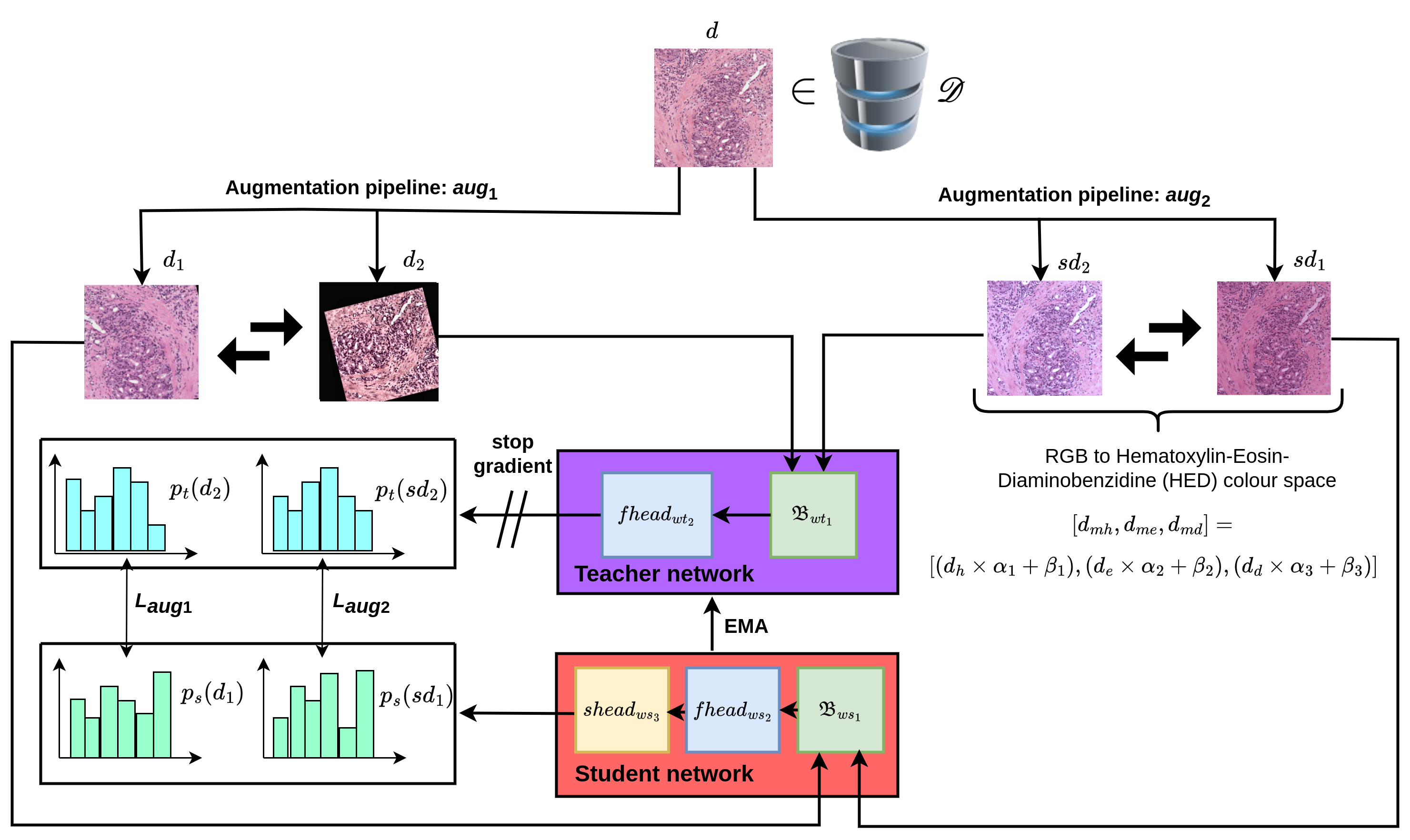}
    \caption{\textbf{Module 2:} Task-specific stain agnostic self-supervised pre-training}
    \label{fig:step2c}
\end{figure*}

Next, a deep model is designed based on a  task-specific \& stain-agnostic SSL framework. It follows the teacher-student knowledge distillation approach \cite{hinton2015distilling}. The aim is to effectively learn the stain-invariant morphological features of GPs using a novel loss function. The training procedure is summarized in Fig. \ref{fig:step2c}. There are two augmentation pipelines involved. The $aug_1$ pipeline involves common strategies such as random color jitter, grayscale conversion, Gaussian blur, horizontal flip, etc. Given an image patch $\mathcal{d} \in \mathscr{D}$, the augmented views $\mathcal{d}_1$  and $\mathcal{d}_2$ are generated from $aug_1$.

In order to learn the stain-invariant characteristics for patch $\mathcal{d}$, two different
stain-augmented views $s\mathcal{d}_1$ and $s\mathcal{d}_2$ are generated using the $aug_2$ pipeline. It simulates a range of realistic H\&E-stained images. First, each patch sample $\mathcal{d}$ is transformed from RGB to Hematoxylin-Eosin-Diaminobenzidine (HED) color space and decomposed into three channels $\mathcal{d}_h$, $\mathcal{d}_e$, $\mathcal{d}_d$ based on color deconvolution \cite{ruifrok2001quantification}. Each of the HED channels is modified as
\begin{equation}
    [\mathcal{d}_{mh},\mathcal{d}_{me},\mathcal{d}_{md}] = [(\mathcal{d}_{h}\times \alpha_1 + \beta_1),(\mathcal{d}_{e}\times \alpha_2 + \beta_2),(\mathcal{d}_{d}\times \alpha_3 + \beta_3)], 
\end{equation}
with random factor $\{\alpha_i\}_{i=1}^3$, bias $\{\beta_i\}_{i=1}^3$, being drawn from a uniform distribution having range [0,1].
The modified channels are then combined, with the sample reconverted to RGB space, to generate a stain-augmented image $s\mathcal{d}_1$. 
The same procedure is repeated by sampling different
values $\{\alpha_i\}_{i=1}^3$, $\{\beta_i\}_{i=1}^3$, to obtain another stain-augmented image $s\mathcal{d}_2$.

The backbone of both teacher and student networks has a similar architecture, with the teacher backbone represented as $\mathfrak{B_{wt_1}}$ and the student backbone denoted by $\mathfrak{B_{ws_1}}$. Here, $wt_1$ and $ws_1$ denote the corresponding model weights. The heads of both networks are asymmetric, to avoid model degeneration \cite{qu2022towards}. Specifically, the head of the student model 

comprises a pair of MLPs as $fhead_{ws_2}$ and $shead_{ws_3}$. The head of the teacher model consists of only one MLP as $fhead_{wt_2}$.
The prediction of the student network, for any patch $img$, is obtained as
\begin{equation}
    \mathcal{p}_s(img) = shead_{ws_3}(fhead_{ws_2}(\mathfrak{B}_{ws_1}(img))).
\label{ssl1_1}
\end{equation}
Analogously, the prediction of the teacher network corresponding to any patch $img$ is represented as
\begin{equation}
    \mathcal{p}_t(img) = fhead_{wt_2}(\mathfrak{B}_{wt_1}(img)).
\label{ssl1_2}
\end{equation}

The parameters of the student network are updated by minimizing the  new loss function
\begin{equation}
    L_{total} = L_{aug_1} + \lambda L_{aug_2},
\label{ltotal}
\end{equation}         
where $\lambda$ is a hyperparameter used to adjust the contribution of $L_{aug_2}$. Here, $\lambda$ was chosen in the range [0.01, 0.03] after several experiments. Since gradients do not flow through the teacher network during training, the parameters are updated through the Exponential Moving Average (EMA) from the student's parameters.

Note that the loss component $L_{aug_1}$, based on the $aug_1$ pipeline, is expressed as
\begin{equation}
    \begin{aligned}
    L_{aug_1} = & \{MSE(L_2norm(\mathcal{p}_s(\mathcal{d}_1)), L_2norm(\mathcal{p}_t(\mathcal{d}_2))) \\
                & + MSE(L_2norm(\mathcal{p}_s(\mathcal{d}_2)), L_2norm(\mathcal{p}_t(\mathcal{d}_1)))\}/2.
    \end{aligned}
\label{ssl1_3}
\end{equation}
Initially, $\mathcal{d}_1$ is passed to the student network and $\mathcal{d}_2$ is transmitted to the teacher network. The corresponding Mean Squared Error ($MSE$) between the
L2-normalized prediction $\mathcal{p}_s(\mathcal{d}_1)$ of the student network [eqn. (\ref{ssl1_1})] and the target output $\mathcal{p}_t(\mathcal{d}_2)$ of the teacher network [eqn. (\ref{ssl1_2})] are computed. To make $L_{aug_1}$ symmetric, the same procedure is repeated by passing $\mathcal{d}_1$ to the teacher network and $\mathcal{d}_2$ to the student network. The corresponding predictions $\mathcal{p}_s(\mathcal{d}_2)$ and $\mathcal{p}_t(\mathcal{d}_1)$ are now calculated from eqns. (\ref{ssl1_1}) and (\ref{ssl1_2}).

Analogously, $s\mathcal{d}_1$ ($s\mathcal{d}_2$) are generated from the $aug_2$ pipeline and passed to the student (teacher) networks, respectively; and vice versa, as before. The corresponding loss component $L_{aug_2}$ is evaluated as in eqn. (\ref{ssl1_3}), but with $s\mathcal{d}_1$ and $s\mathcal{d}_2$ instead of $\mathcal{d}_1$ and $\mathcal{d}_2$, respectively. 

\subsection*{Module 3: Fine-tuning for ISUP grading}

\begin{figure*}[!ht]
    \centering
    \includegraphics[width=0.9\textwidth]{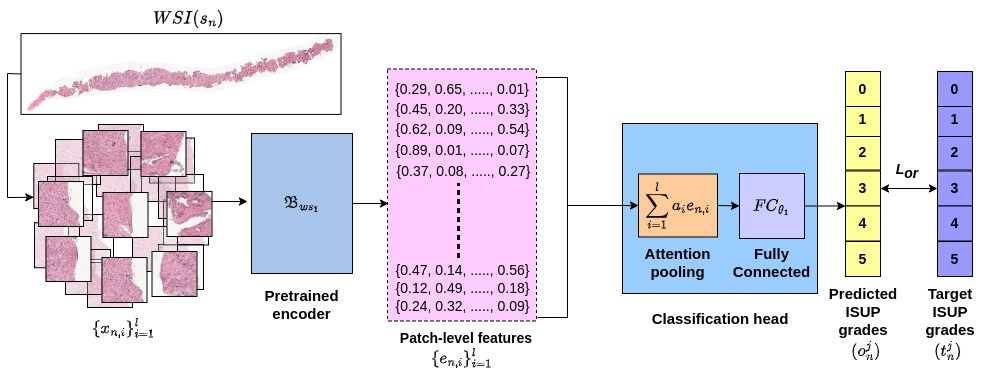}
    \caption{\textbf{Module 3:} Fine-tuning for ISUP grading}
    \label{fig:step3c}
\end{figure*}

The backbone of the pre-trained student network is adapted for the task of ISUP grading, by
posing it as an ordinal regression problem. The training framework is depicted in Fig. \ref{fig:step3c}. Each label or ISUP grade $z_{n,isup}$ is converted to the corresponding target vector
$\mathbf{[t_n^j]_{j=1}^5}$, where the sum of the five elements is equal to the corresponding ISUP grade. For example, ISUP grade 0 is mapped to [0, 0, 0, 0, 0], ISUP grade 2 is mapped to [1, 1, 0, 0, 0], etc. This kind of label encoding, as opposed to the one-hot encoding of traditional classification models, ensures that the difference between classes follows an ordinal scale. For example, the difference between ISUP grades 2 and 5 is greater than that between grades 2 and 3 when such encoding is used. It helps to ascertain that errors made by the deep model are close to the actual grade. This is highly desirable in grading medical images.

The backbone of the pre-trained student network $\mathfrak{B_{ws_1}}$ uses a custom classification head during training for ISUP grading. It comprises an efficient attention-based MIL pooling operator for improved interpretability, followed by a fully connected layer. 
The convolutional part of $\mathfrak{B_{ws_1}}$ generates low-dimensional patch-level embedding 
\begin{equation}
    \{e_{n,i}\}_{i=1}^l = \mathfrak{B_{ws_1}}(\{x_{n,i}\}_{i=1}^l),
\label{lisup_1}
\end{equation}
with $l$ denoting the number of patches from each WSI. Patches are concatenated using the attention-based MIL pooling operator, evaluated as  
\begin{equation}
    a_i = \frac{exp\{W_{\theta_3}^T tanh(V_{\theta_2}e_{n,i}^T)\}}{\sum_{j=1}^{l} 
    exp \{W_{\theta_3}tanh(V_{\theta_2}e_{n,j}^T)\}},
\label{lisup_3}
\end{equation}
where $V_{\theta_2}$, $W_{\theta_3}$ are two fully connected layers, $tanh$ is the activation function, $T$ denotes transpose operation, and $exp$ is the exponentiation operation. Here, $a_i$ provides each embedding $e_{n,i}$ with the context about its role in contributing to the WSI label.
The fully connected layer $FC_{\theta_1}$ predicts the output vector 
\begin{equation}
        \mathbf{[o_n^j]_{j=1}^5} = FC_{\theta_1}(\sum_{i=1}^{l}a_i, e_{n,i}),
\label{lisup_2}
\end{equation}
with the training loss being computed as
\begin{equation}
    L_{or} = BCE(\mathbf{[o_n^j]_{j=1}^5}, \mathbf{[t_n^j]_{j=1}^5}).
\label{lisup_4}
\end{equation}
\section{Implementation and Experimental Details}
\label{sec:exp_setup}

 This section describes the publicly available prostate WSI datasets used along with implementation details of the algorithms. This is followed by experimental results obtained for the detection and grading of prostate cancer. It includes ablation studies, as well as comparison with related state-of-the-art (SOTA) methods.
 
\subsection{Datasets used}

The model was trained with the largest multicenter prostate biopsy WSI dataset from the Kaggle Prostate cANcer graDe Assessment (PANDA) Challenge \cite{kagglepanda}. The testing phase was carried out on data originating from two different centers, {\it i.e.} the PANDA and SICAP datasets \cite{silva2020going}. PANDA consists of $10,616$ H\&E-stained digitized biopsies, whose primary and secondary GG were labeled by expert pathologists. Considering the trade-off between the magnification details and the associated computational complexity, we considered the intermediate-resolution (5X) images for our experiments. The SICAP database, used for testing, consists of 155 WSI with global primary and secondary GG annotated by expert pathologists. As the images are available at 10X magnification, they were downsampled to 5X resolution to maintain uniformity. 

The distribution of WSIs, used for training, validation, and testing with respect to ISUP grades, is provided in Table \ref{tab1}. ISUP 0 images are not required in Module 1. Around 4000 patches of each class, from the training partition of WSIs, are used for SSL pre-training of Module 2. Finally, Module 3 works on the data splits summarized in the table. Annotations of all GP are available in case of SICAP, whereas only one center's annotations (covering 49\% of the data) are provided for PANDA.

\begin{table}[!ht]
\caption{Distribution of prostate biopsy WSI used in learning}
\centering
\begin{tabular}{|l|l|l|l|l|}
\hline
\multicolumn{1}{|c|}{ISUP} & \multicolumn{1}{c|}{\begin{tabular}[c]{@{}c@{}}Training\\ (PANDA)\end{tabular}} & \multicolumn{1}{c|}{\begin{tabular}[c]{@{}c@{}}Validation \\ (PANDA)\end{tabular}} & \multicolumn{1}{c|}{\begin{tabular}[c]{@{}c@{}}Testing\\ (PANDA)\end{tabular}} & \multicolumn{1}{c|}{\begin{tabular}[c]{@{}c@{}}Testing\\ (SICAP)\end{tabular}} \\ \hline
0                          & 922                                                                          & 166                                                                         & 272                                                                         & 36                                                                          \\ 
1                          & 922                                                                          & 166                                                                         & 271                                                                         & 14                                                                          \\ 
2                          & 921                                                                          & 166                                                                         & 272                                                                         & 22                                                                          \\ 
3                          & 922                                                                          & 166                                                                         & 273                                                                         & 23                                                                          \\ 
4                          & 904                                                                          & 116                                                                         & 255                                                                         & 18                                                                          \\ 
5                          & 913                                                                          & 100                                                                         & 253                                                                         & 42                                                                          \\ \hline
\end{tabular}
\label{tab1}
\end{table}

\subsection{Implementation details}

The lightweight EfficientNet-B1 \cite{tan2019efficientnet} was used as the backbone for all deep models used here, with patches of size $256\times256$ as input. Patches from each slide were first sorted, based on average pixel intensity, in ascending order. The top 36 patches, having the lowest average pixel intensity, were considered from each slide for training in Modules 1 and 3. 
Runtime augmentation was used at patch level during training in Modules 1 \& 3, using affine transforms such as random flipping, rotation, scaling; along with color augmentations such as random changes in brightness, contrast, saturation, and hue. Adam optimizer was used in all three modules, along with the cosine annealing learning rate scheduler, involving an initial learning rate of $3e^{-4}$. Considering batch sizes of 8, 40, 8, in Modules 1, 2, 3, respectively, the corresponding deep models were trained for 35, 50, and 30 epochs, respectively. The implementations were made in the Pytorch framework (version 1.13.1) using Python (version 3.9) with a dedicated GPU (NVIDIA RTX A6000 of capacity 48 GB).

\subsection{Results: ablations and comparisons}

The metrics used to evaluate the different modules of $TSOR$ include the $F1$ score, the Area under the Receiver Operating Characteristic (ROC) curve (AUC), Accuracy, and Cohen's quadratic Kappa \cite{silva2021self}. The Cohen quadratic kappa is a statistical metric used to measure the agreement between two raters, while accounting for the possibility of agreement occurring by chance. It is a weighted version of Cohen's Kappa, where disagreements get penalized more when they are larger. It is expressed as
\begin{equation}
\label{metric:kappa}
    Kappa = 1 - \frac{\sum_{i,j} W_{ij} O_{ij}}{\sum_{i,j} W_{ij} E_{ij}},
\end{equation}
where $O_{ij}$ is the observed confusion matrix, $E_{ij}$ is calculated as the outer product between the actual histogram vector of outcomes and the predicted histogram vector, $W_{ij}$ is the quadratic weight calculated as $W_{ij} = \left( \frac{i - j}{k - 1} \right)^2$. Here, $i$ and $j$ are categories, and $k$ is the number of possible categories.

The deep model, with the best values (w.r.t. $F1$ score, accuracy) as obtained on the validation set in Module 1, was used to label patches for the SSL pre-training step. The network generated in Module 2, with the lowest validation loss, was used as the backbone for Module 3. 
The efficacy of the proposed framework $TSOR$ is demonstrated for both prostate cancer detection and grading tasks. Although detection was aimed at determining whether a prostate WSI is benign (ISUP 0) or malignant (ISUP 1-5), the grading task classified the WSI into one of the six ISUP grades (0-5).

\subsubsection{Ablations} The impact of the task-specific and stain-agnostic SSL pre-training, as well as ordinal regression in case of ISUP grading, was evaluated through extensive ablation studies for the detection and grading in test data sets. The best results are marked in bold.

\paragraph{Detection} Table \ref{tab4} shows the ablation study for prostate cancer detection, over both test datasets, w.r.t. the performance metrics. Specifically, we evaluated Module 3 of our framework, pre-trained using task-specific SSL [$L_{aug_1}$ \{eqn. (\ref{ssl1_3})\}], task-specific \& stain-agnostic SSL [$L_{total}$ \{eqn. (\ref{ltotal})\}], and without SSL pre-training (ImageNet pre-trained). A clear improvement can be observed when Module 3 is pre-trained using task-specific \& stain-agnostic SSL, as compared to the rest. The observation corroborates that our $TSOR$ framework, pre-trained using the same approach, performs consistently across both datasets; thereby, justifying its robustness. 

\begin{table}[!ht]
\caption{Performance of Module 3 of $TSOR$, on the test datasets, in terms of different metrics}
\setlength{\tabcolsep}{3pt}
\centering
\begin{tabular}{|p{55pt}|p{35pt}| p{28pt}| p{28pt}| p{35pt}| p{28pt} | p{28pt}| }
 \hline
 Data $\rightarrow$ & \multicolumn{3}{c|}{SICAP} & \multicolumn{3}{c|}{PANDA} \\
 \hline
 Method                               & Accuracy     & F1 score   & AUC    & Accuracy     & F1 score   & AUC  \\
 \hline
 Without SSL                          & 0.8129       & 0.7600     & 0.9939 & 0.9167      & 0.9031     & 0.9675 \\
 \hline
 Task-specific SSL                    & 0.8968       & 0.8853     & 0.9988 & 0.9336      & 0.9342     & 0.9750 \\
 \hline
 Task-specific \& stain-agnostic SSL     & \textbf{0.9097}       & \textbf{0.9013}     & \textbf{1.0} &\textbf{0.9386}      & \textbf{0.9399}     & \textbf{0.9762}  \\
 \hline
\end{tabular}
\label{tab4}
\end{table}

A high $AUC$ value across both datasets indicates that our $TSOR$ is good at distinguishing between the positive and negative classes; thus, achieving a good balance between true positives and false positives. This is highly desirable in an automated diagnostic system.

\paragraph{Grading} The ablation study for prostate cancer grading, by Kappa score, is shown in Table \ref{ablation_grading}. We evaluated the impact of pre-training using the task-specific and 
stain-agnostic SSL framework, along with fine-tuning involving ordinal-regression (OR), on the ISUP grading performance of Module 3. Here, fine-tuning ``without OR" implies training using the more conventional one-hot encoding with Cross Entropy loss. It is evident from the last column that our $TSOR$ performed consistently well on test data. This highlights the robustness of $TSOR$ to the domain shift caused by variations in stain color. 

\begin{figure} [!ht]
\begin{center}
\begin{tabular}{cc} 

\includegraphics[width=0.48\linewidth]{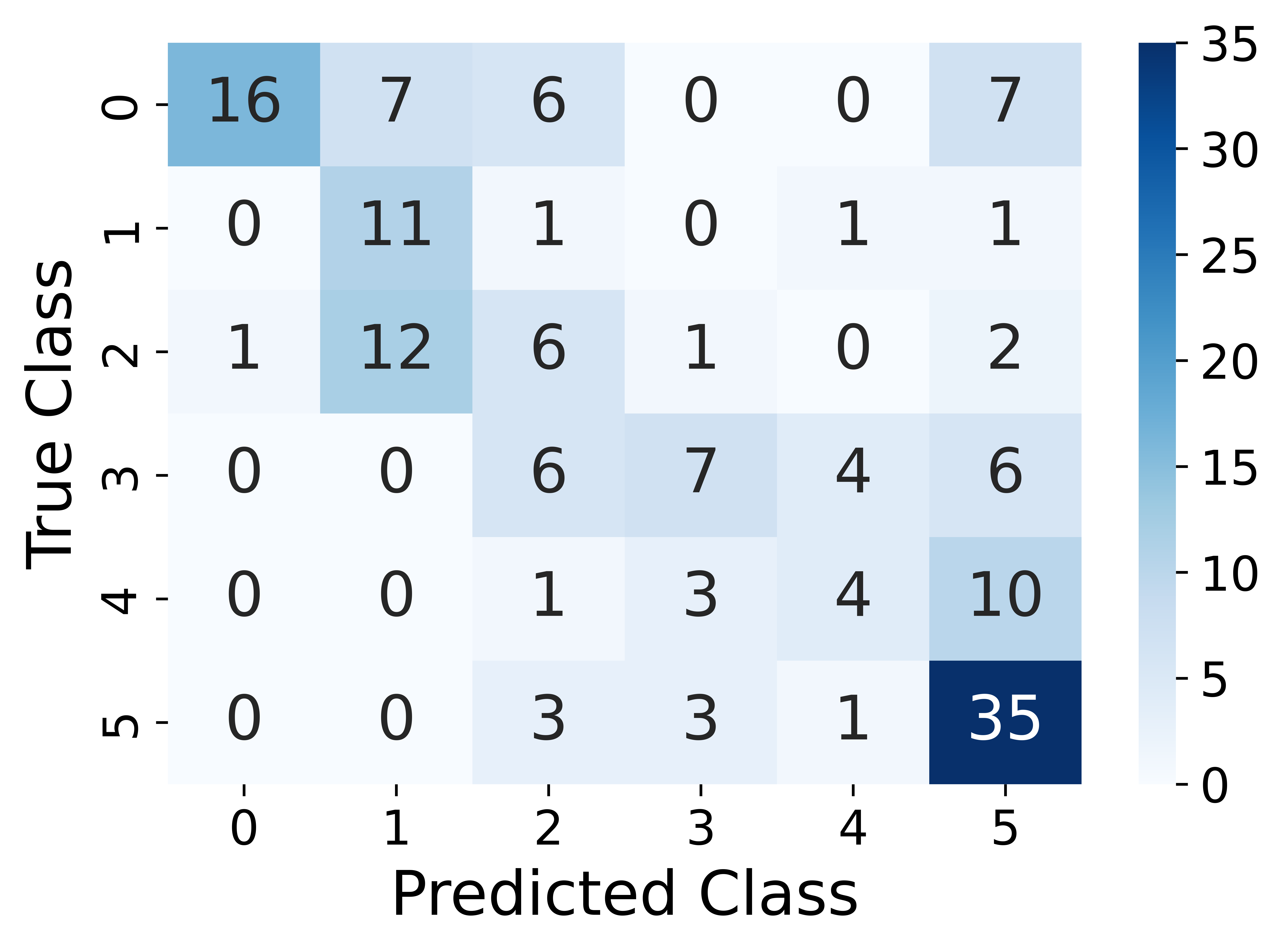}  & 
\includegraphics[width=0.48\linewidth]{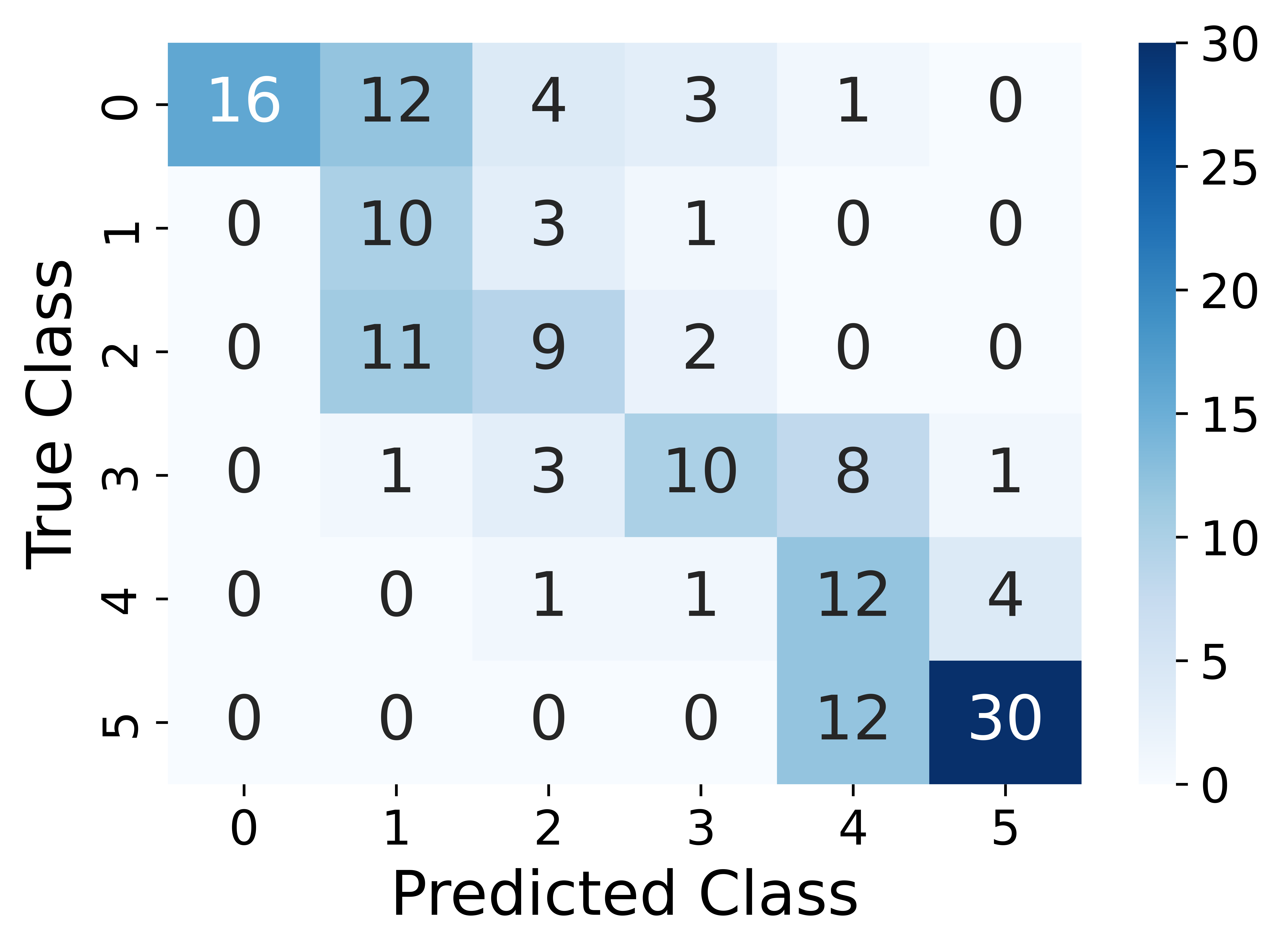}\\
(a) & (b) \\
\includegraphics[width=0.48\linewidth]{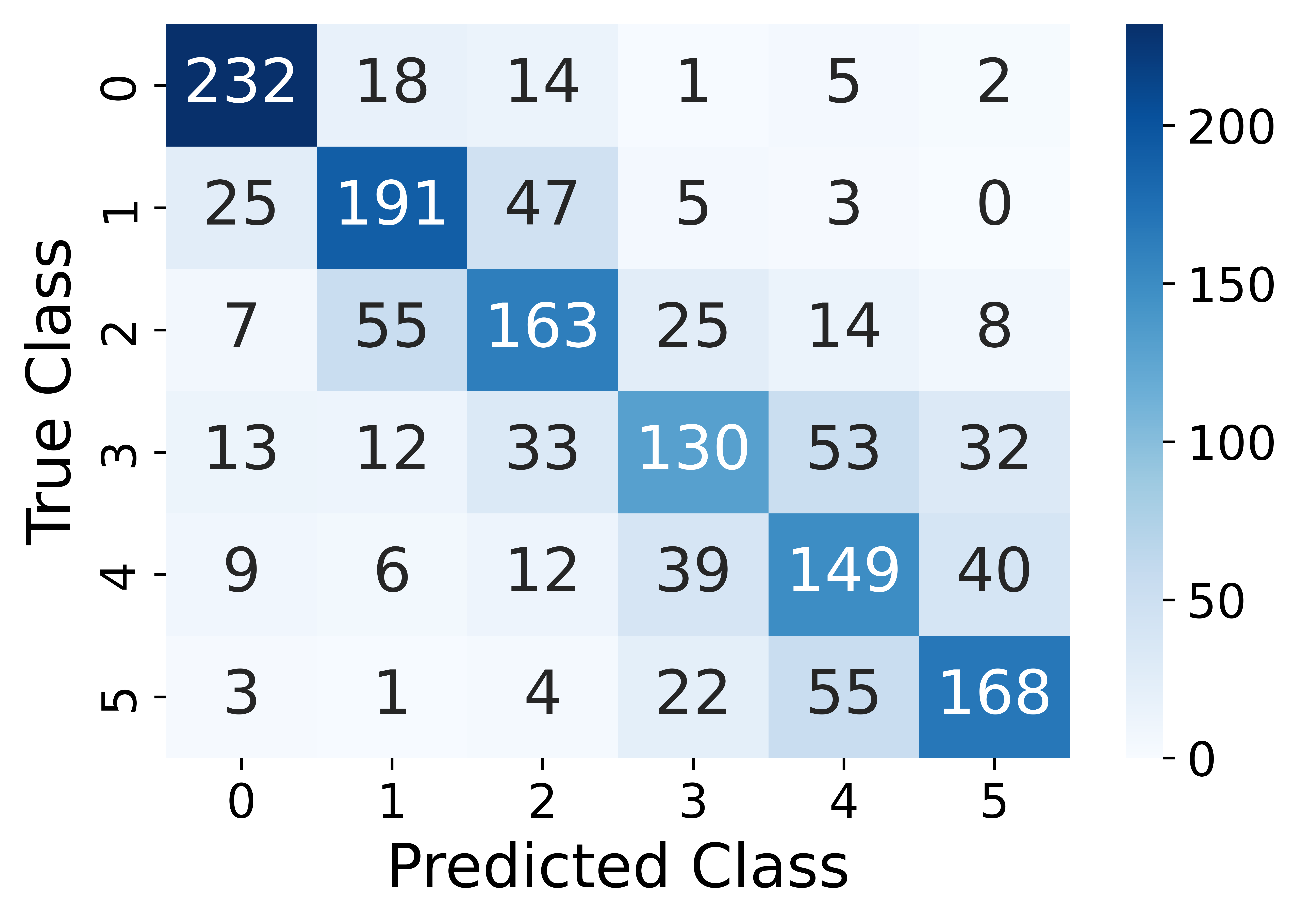}  & 
\includegraphics[width=0.48\linewidth]{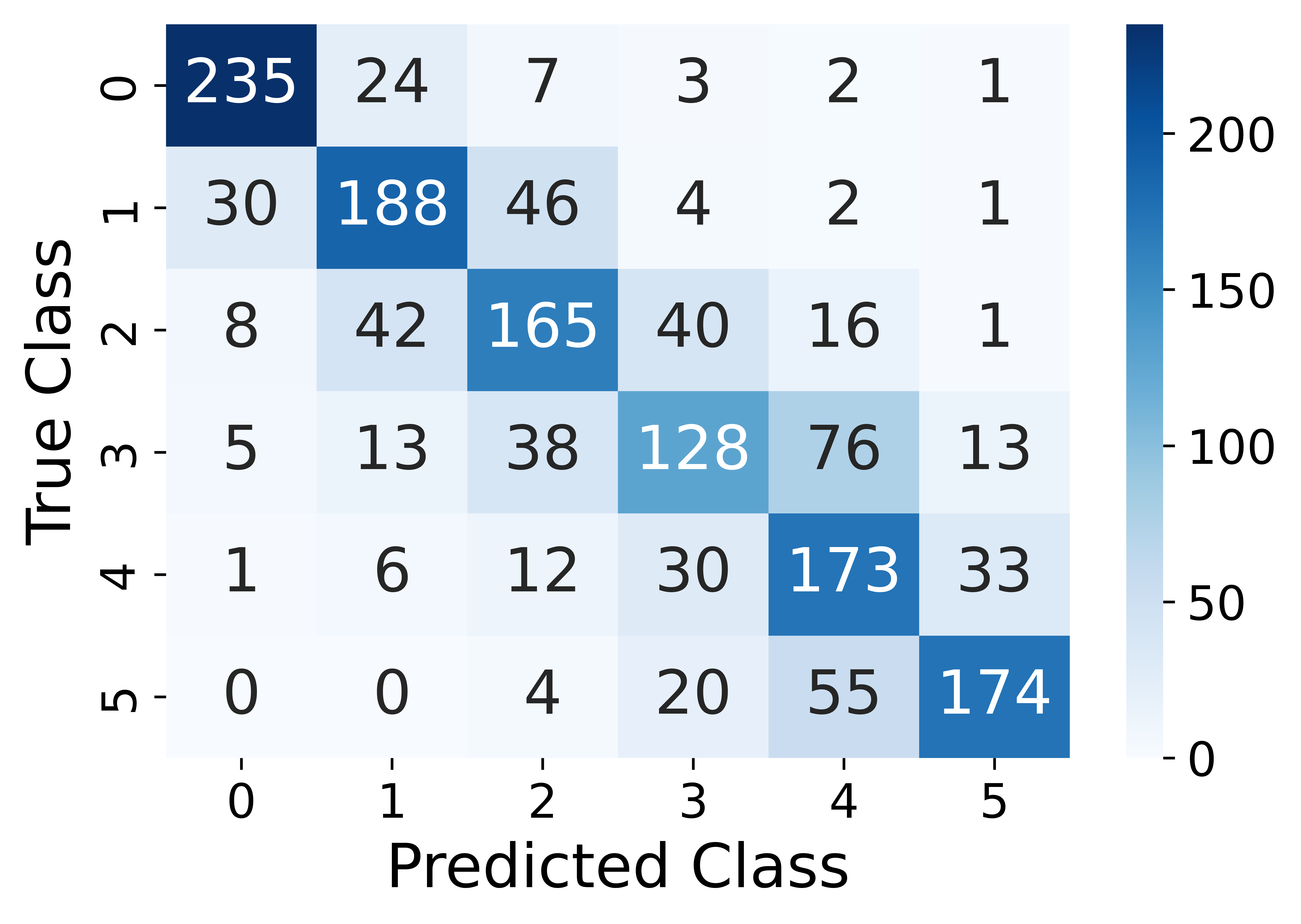}\\
(c) & (d) \\
\end{tabular}
\end{center}
\caption{Confusion matrices generated for ISUP grading, using Module 3 of $TSOR$: \\ Task-specific \& 
stain-agnostic SSL with fine-tuning without OR on (a) SICAP, (c) PANDA; and \\ Task-specific \& stain-agnostic SSL with fine-tuning and OR, on (b) SICAP, (d) PANDA} 
\label{grading}
\end{figure}

The confusion matrices in Fig. \ref{grading} show that the prediction errors occur more in grades adjacent to the desired (true) one, when our OR-based approach (in lieu of the more conventional one-hot encoding scheme) is used for fine-tuning. Thus in parts (b) and (d) the model avoids extreme misclassification in the top-right and bottom-left corners. This suggests improved class separation, leading to fewer severe prediction errors [compared to those in parts (a) and (c) of the figure]. For example, consider the incorrect predictions for true grade 5 in Fig. \ref{grading}(a). These correspond to 3, 3, 1, for grades 2, 3, 4, respectively. Upon incorporating OR, the confusion shifts entirely to the adjacent predicted grade 4 in part (b). Besides, the number of predictions for grades 0, 1, 2 getting misclassified as grade 5 in part (a), become zero in part (b) of the figure. Comparing parts (c) and (d) of the figure demonstrates a reduction in the number of grade 3, 4, 5 images getting misclassified as grade 0 in part (d), along with a simultaneous increase in the number of 
mis-predictions being made only in positions adjacent to the actual grade. It corroborates the effectiveness of employing ordinal regression, during fine-tuning, for ISUP grading.

\begin{table}[!ht]
\caption{Grading performance of $TSOR$, on the test datasets, in terms of Kappa}
\setlength{\tabcolsep}{3pt}
\centering
\begin{tabular}{|p{135pt}|p{30pt}| p{30pt}|p{30pt}|}
 \hline
 Dataset $\rightarrow$ & SICAP & PANDA & Average \\
 \hline
 Without SSL \& Without OR                      & 0.7208                &  0.8250  & 0.7729 \\
 \hline
 Without SSL \& OR                              & 0.8129                &  0.8542  & 0.8335    \\
 \hline
 Task-specific SSL \& OR                        & 0.8612                &  0.8705  & 0.8658  \\
 \hline
 Task-specific \& stain-agnostic SSL \& OR       & 0.8736               &  0.8843  & \textbf{0.8789}\\
 \hline
\end{tabular}
\label{ablation_grading}
\end{table}

\subsubsection{Comparative study}

Here we summarize the comparisons for the detection and ISUP grading of prostate cancer, with related SOTA from the literature.

\begin{figure}[!ht]
    \centering
    \includegraphics[width=0.45\textwidth]{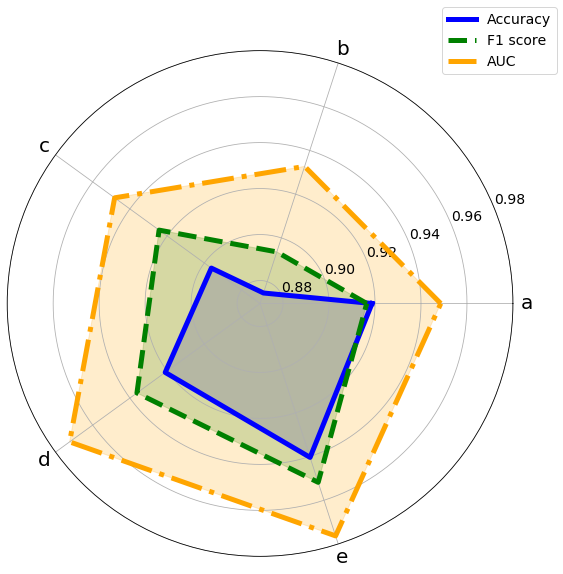}
    \caption{Comparative performance of proposed (e) $TSOR$, in terms of Radar plot, with SOTA including (a) TMIL \cite{shao2021transmil}, (b) DSMIL \cite{li2021dual}, (c) CLAM \cite{lu2021data}, (d) SCL-WC \cite{wang2022scl}, for the detection of prostate cancer on PANDA test set}
    \label{fig:comparison1}
\end{figure}

\paragraph{Detection} The performance of our $TSOR$ was compared with
CLAM \cite{lu2021data}, DSMIL \cite{li2021dual}, TMIL \cite{shao2021transmil}, SCL-WC \cite{wang2022scl}, using Accuracy, $F1$ score, $AUC$ metrics on the PANDA test set. The results are illustrated in the Radar plot in Fig. \ref{fig:comparison1}. It is observed that the $TSOR$ model consistently outperformed the SOTA. Note that Ref. \cite{shao2021transmil,lu2021data} used  Imagenet pre-trained ResNet50 to extract patch-level features. Ref. \cite{li2021dual, wang2022scl} employed a task-agnostic, computationally expensive SSL framework which was pre-trained on a large database of different histopathology images. In addition, no mechanism was incorporated to handle the stain variation. Thus, $TSOR$, pre-trained using a stain-agnostic and task-specific SSL framework, performed consistently better compared to these methods.

\paragraph{Grading} The ISUP grading performance of $TSOR$ was compared with that of the SOTA methods \cite{nagpal2019development,silva2020going,nirthika2020loss,yang2021multi,silva2021self,yang2023devil}. 
Patch-level annotations were used in Ref. \cite{nagpal2019development,silva2020going} to perform supervised training to extract the corresponding patch-level features. Since patch labels were inferred from WSIs \cite{silva2021self}, the patch-level training becomes noisy and is reflected in its ISUP grading. While a task-agnostic SSL framework was employed \cite{yang2023devil} for patch-level feature extraction, the  Imagenet pre-trained models were used in Ref. \cite{nirthika2020loss, yang2021multi}. None of these approaches focused on handling the issues that occur due to stain-variation in WSIs. 

Our ISUP grading framework $TSOR$, based on task-specific and stain-agnostic SSL, is able to outperform these methods without using patch-level annotations. Furthermore, the ordinal regression factor during the fine-tuning of the pre-trained model helps to improve the Kappa score. The results on the PANDA dataset are depicted in terms of Kappa, in Table \ref{comp_panda}. Kappa scores of 0.52 by \cite{nagpal2019development}, 0.50 by \cite{silva2020going}, and 0.87 by our proposed $TSOR$, on SICAP data, further demonstrate that $TSOR$ is able to generalize better. 


\begin{table}[!ht]
\caption{Comparative study of ISUP grading on the PANDA data}
\label{table}
\setlength{\tabcolsep}{3pt}
\centering
\begin{tabular}{|p{240pt}|p{40pt}|}
 \hline
 Method                                                                          & Kappa score   \\
 \hline
 Self-learning system (Features + Average + MLP)\cite{silva2021self}             & 0.8245  \\
 \hline
 Self-learning system (Features + Average + kNN)\cite{silva2021self}             & 0.7927  \\
 \hline
 GG\% + kNN \cite{nagpal2019development}                                         & 0.8152   \\
 \hline
 GG\% + MLP \cite{silva2020going}                                                & 0.8229    \\
 \hline
 ISMIL  \cite{yang2023devil}                                                     & 0.8600    \\
 \hline
 Bag-based MIL \cite{nirthika2020loss}                                           & 0.8120    \\
 \hline
 Multichannel and Multispatial Attention  CNN  \cite{yang2021multi}              & 0.8503    \\
 \hline                                                 
 $TSOR$                                                                 & \textbf{0.8843}     \\
 \hline
 \end{tabular}
\label{comp_panda}
\end{table}

\subsubsection{Explainability}

\begin{figure*}[!ht]
    \centering
    \includegraphics[width=0.8\textwidth]{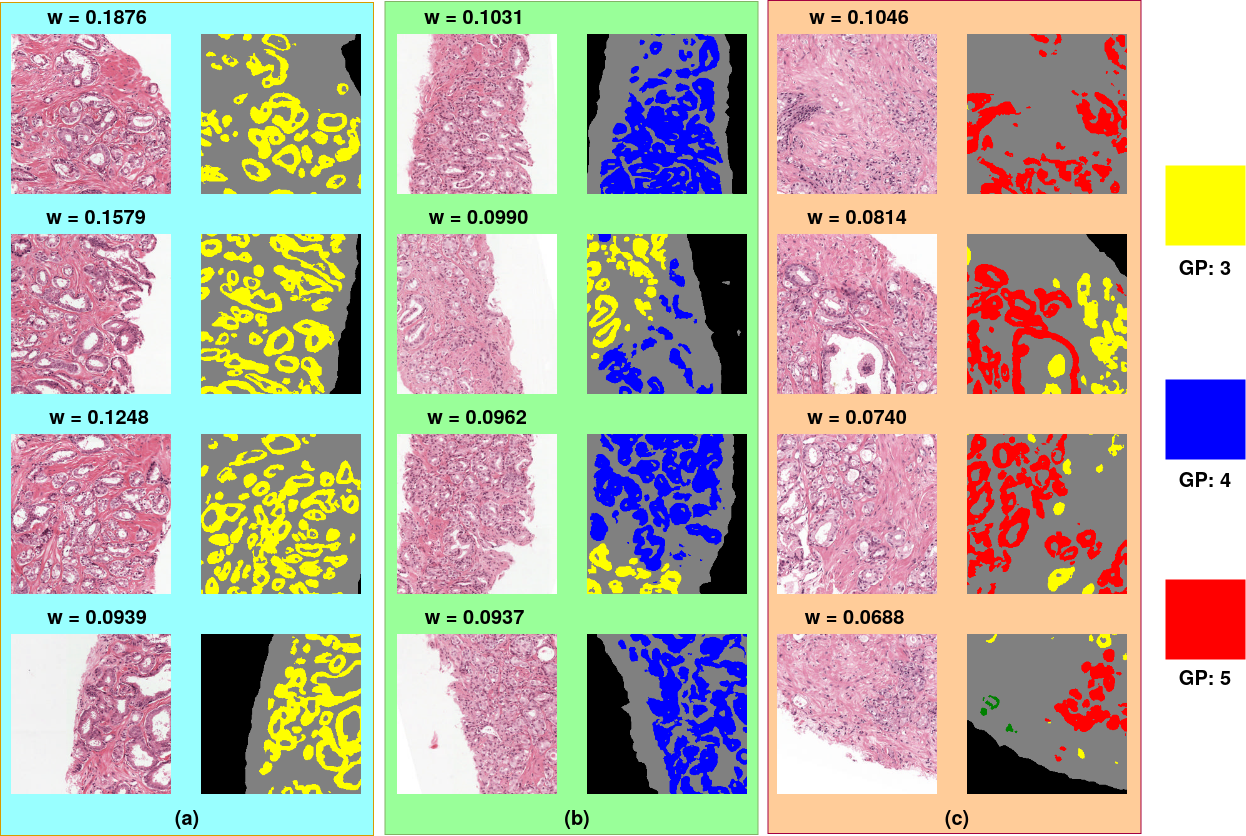}
    \caption{Sample patches of high diagnostic importance, in terms of the weights [eqn. (\ref{lisup_3})] 
     predicted by $TSOR$, for three images from test data having ISUP grades (a) 1, (b) 3, and (c) 4. Each patch has its corresponding mask to its immediate right.}
    \label{visual_results}
\end{figure*}

To provide confidence to pathologists, an automated diagnostic system should offer explainability regarding the interpretation of a given WSI of prostate cancer. The attention module of our $TSOR$ helps identify the contribution of different patches to the final WSI grading, by assigning corresponding weights. Patches receiving greater weight should indicate regions of high diagnostic importance. Fig. \ref{visual_results} shows the patches of some of the annotated cancerous WSIs, that were weighted high. It is evident that the patches that are assigned a greater weightage by the model are actually relevant regions, as validated from the corresponding masks provided by expert pathologists. This shows that the decision provided by $TSOR$ is interpretable for the model to be considered reliable by human end users (here, pathologists).

\section{Conclusions and Discussion}
\label{sec:conclusion}

An efficient SSL-based prostate cancer detection and ISUP grading framework was developed for the WSI of prostate biopsies, using only slide-level labels. Cancerous patterns in prostate cancer are highly heterogeneous and diverse, with stain variation between centers further affecting the generalizability of the models. In this context, it is essential to initially learn the patch-level features of Gleason patterns for effective WSI-level grading. Previous methods in the literature for prostate cancer detection or grading used supervised 
patch-level classifiers trained using expensive patch-level annotations \cite{nagpal2019development, silva2020going}, Imagenet pre-trained models with huge domain discrepancy \cite{shao2021transmil,lu2021data, nirthika2020loss, yang2021multi} or computationally expensive task-agnostic SSL frameworks \cite{yang2023devil, li2021dual, wang2022scl} to extract patch-level features. In addition, none of these methods focused on handling the stain variation across centers. In contrast, a novel task-specific and stain-agnostic SSL framework was designed here with a new loss term based on stain augmentation. 

First, a well-balanced patch-level dataset was created using pseudo-labeling within an efficient hybrid MIL framework using only slide-level labels. This ensured effective task-specific self-supervised pre-training, to learn the stain-agnostic patch-level morphological features which are critical for effective WSI grading. Moreover, the proposed procedure for labeling patches using only slide-level labels and creating a well-balanced 
patch-level dataset can also be employed in other tasks with similar requirements.

Finally, the fine-tuning of the pre-trained network was performed in terms of ordinal regression, in contrast to the common approach of using one-hot encoded labels with Cross Entropy loss. This helped $TSOR$ to make fewer severe prediction errors (as observed from the confusion matrices); thereby, making it better suited for medical image grading. Also, the use of nonparametric classifiers like $k$-NN \cite{nagpal2019development, silva2021self} for final grading has the inherent problems of sensitivity to noise, choice of $k$ and high dimensionality. The consistent performance of $TSOR$ across different datasets while outperforming the state-of-the-art algorithms by considerable margin, as evidenced from the experimental results, indicated the effectiveness of our pre-trained neural network-based method for ISUP grading. In contrast, the comparative study suggested a severe drop in performance of the other methods when tested on external datasets for ISUP grading. The proposed patch-level 
task-specific SSL pre-training also helped our framework to achieve good performance in prostate cancer detection, while remaining unaffected by the inherent class-imbalance problem among cancerous and non-cancerous classes during detection. A high AUC value across datasets suggested that our framework is effective in identifying true positives while minimizing false positives. Thus, the $TSOR$ framework achieved performance comparable to pathologists in both prostate cancer detection and grading tasks, using only intermediate resolution images that involve only slide-level labels readily available from clinical records. Moreover, the interpretability associated with the results justifies the trustworthiness of our framework.

As prostate cancer is one of the leading causes of cancer-related deaths among men worldwide, the clinical relevance of this research is significant. The ISUP grading procedure by pathologists is time-consuming and is highly prone to intra and inter-observer variability. Thus, the ability to provide an accurate and interpretable automated grading system can assist pathologists make more consistent and timely diagnoses with high confidence; particularly, in regions with limited access to specialized medical expertise. The high Kappa score, achieved by our framework across different datasets, suggests a strong agreement between the predicted and actual labels with the errors being mostly adjacent to the actual grade. Thus, this automated system, through reduction in workload on healthcare professionals and minimization of diagnostic variability, has the potential to improve patient outcomes and enhance the overall efficiency of treatment. 

Some of the limitations of the study point to potential future scopes. These include the requirement of GPUs with considerable memory and the limited availability of a substantial number of publicly available graded prostate cancer datasets for experimentation. Although this method gives good performance by empirically selecting a constant number of patches from each WSI during training, it would be interesting to consider using other patch selection strategies with fewer WSIs for training. It would also be interesting to explore other SSL paradigms for patch-level pre-training. Aggregating patch-level features, using spatial-context-aware architectures like Vision Transformers (ViT) and Graph Convolution Networks (GCN) would also be an interesting research direction. Further, one can attempt to find the primary and secondary GG of each WSI, along with the ISUP grade. In addition, other explainability techniques can be explored in such MIL scenarios.

\section*{Acknowledgments}
This research was supported by the J. C. Bose National Fellowship, grant no. JCB/2020/000033 of S. Mitra.
\bibliographystyle{unsrt}  
\bibliography{references}

\end{document}